\documentclass[10pt,twocolumn,letterpaper]{article}

\usepackage[dvipsnames]{xcolor}
\usepackage[T1]{fontenc}
\usepackage{lmodern}
\usepackage[format=plain,labelformat=simple,labelsep=period,font=small,compatibility=false]{caption}
\usepackage[font=footnotesize,skip=3pt,subrefformat=parens]{subcaption}
\usepackage[pagebackref,breaklinks,colorlinks]{hyperref}
\usepackage{amsmath}
\usepackage[capitalize]{cleveref}
\crefname{section}{Sec.}{Secs.}
\Crefname{section}{Section}{Sections}
\Crefname{table}{Table}{Tables}
\crefname{table}{Tab.}{Tabs.}
\usepackage{etoolbox}
\AtBeginEnvironment{tabular}{\small}
\usepackage[numbers,sort&compress]{natbib}
\usepackage[utf8]{inputenc} 
\usepackage[T1]{fontenc}    
\usepackage{hyperref}       
\usepackage{url}            
\usepackage{booktabs}       
\usepackage{amsfonts}       
\usepackage{nicefrac}       
\usepackage{microtype}      
\usepackage{xcolor}         
\usepackage{graphicx}       
\usepackage{amsmath}
\usepackage{placeins}
\usepackage{enumitem}
\graphicspath{{img/}}

\newcommand{\censored}[1]{{#1}}

\title {On the power of data augmentation for head pose estimation}
\date{}

\author{%
  Michael Welter \\
  Independent Researcher \\
  {\tt\small michael@welter-4d.de} \\
}

\begin{document}

\maketitle

\begin{abstract}
Deep learning has been impressively successful in the last decade in predicting
human head poses from monocular images. However, for in-the-wild inputs the research
community relies predominantly on a single training set, 300W-LP, of semisynthetic
nature without many alternatives. This paper focuses on gradual extension and 
improvement of the data to explore the performance achievable with 
augmentation and synthesis strategies further. Modeling-wise a novel multitask
head/loss design which includes uncertainty estimation is proposed. Overall, the
thus obtained models are small, efficient, suitable for full 6 DoF pose estimation,
and exhibit very competitive accuracy.
\end{abstract}

\renewcommand{\deg}{^{\circ}}

\section{Introduction}
\label{sec:intro}

Human head pose estimation (HPE) is an important task in many application e.g.
in the automotive sector or in entertainment products. The concrete problem addressed
by this paper is monocular head pose estimation for in-the-wild scenarios, i.e. 
from given a facial crop a computer vision system must estimate
the face orientation. Closely related
is the task of face alignment where a mathematical description of a face is
inferred, e.g. as landmarks or full 3D reconstruction. These tasks are
particularly challenging due to the diversity and non-rigid nature of faces.
However, in recent years deep learning methods have been very successful at it
\cite{10.1007/s42979-023-01796-z}. In HPE, the current state of the art achieves
errors of ca. $3$ (MAE) to $5\deg$ (geodesic)
\cite{COBO2024110263,DBLP:conf/cvpr/LiWCKT23}.

The main interest behind this work lies in devising a simple, efficient and
effective HPE model. Because the margin for model size and complexity is therefore
limited, the choice of the training data is a knob left to turn freely. In this
regard the goal this paper is exploration of this path toward further improvements
of HPE models. Extension and combination of existing datasets, as well as 
augmentation is the means to this end which might also serve the community 
in the future with custom datasets. However, first we shall review related
literature.

For in-the-wild HPE we can recognize two datasets as de facto standard: AFLW2000-3D and
300W-LP \cite{DBLP:conf/cvpr/ZhuLLSL16}. AFLW2000-3D, serving as test set,
consists of 2000 images labeled with 6 DoF poses, parameters of deformable 
facial 3D model, as well as landmarks. The images are challenging due to occlusion,
extreme poses and varying illumination. 300W-LP, commonly used as training data, 
similarly consists ca. 61 thousand labeled images. It was
constructed by augmenting a smaller set of faces with out-of-plane rotations. 
Some authors \cite{DBLP:conf/eccv/GuoZYYLL20} chose to expand this data
further with fine-grained movements, or employ
a face swapping augmentation \cite{DBLP:conf/3dim/WuXN21}. However, the impact of
this was not considered in isolation.

In contrast, the Biwi dataset \cite{DBLP:journals/ijcv/FanelliDGFG13}, 
popular as another benchmark, comprises 24 sequences including extreme poses 
of 20 subjects 
in a laboratory environment.
Interestingly, it is used in recent works \cite{DBLP:conf/cvpr/GuYMK17,DBLP:conf/iccv/KuhnkeO19,DBLP:journals/tbbis/KuhnkeO23} 
which train on fully synthetic images from the SynHead dataset 
\cite{DBLP:conf/cvpr/GuYMK17}. They are however limited to the lab setting and
a low number of individuals.

Regarding mathematical models for HPE, a baseline may consist of a learned
feature extractor, such as a convolutional neural network and final linear
layers which, after some simple transformation, output orientation, position,
size and so on. Such a network could be trained with losses such as L1 or L2
penalizing the errors from the known ground truths. Various rotation
representations have been introduced \cite{DBLP:journals/tmm/HsuWWWL19,
DBLP:conf/icip/HempelAA22,COBO2024110263}. More
sophisticated model architectures and loss calculations
are proposed in \cite{DBLP:conf/3dim/WuXN21, DBLP:conf/cvpr/LiWCKT23,
DBLP:journals/tip/RuanZWWW21}. The authors of \cite{DBLP:conf/eccv/GuoZYYLL20}
devised an algorithm switching between different losses dynamically.
Multitask networks leveraging synergies between 3D
HPE and 2D landmark prediction were considered in \cite{DBLP:journals/pami/ValleBB21}. 
Competitive HPE results could also be achieved by suitably shaped, yet relatively
simple loss functions \cite{COBO2024110263}.

Uncertainty estimation, while rarely addressed in the face analysis
community \cite{DBLP:journals/tmm/LiuFZLLW22},
is potentially useful for filtering and outlier rejection. However, exploration 
downstream applications is beyond the scope of this paper. Ultimately, uncertainty
estimation is included because it boosted the model's accuracy. For simplicity, 
only aleathoric (data) uncertainty is modeled. This is implemented by outputting 
parameters for an assumed probability density of the data and taking its negative
log likelihood as loss \cite{DBLP:conf/nips/Lakshminarayanan17,DBLP:conf/nips/KendallG17}.
In the space of rotations $SO(3)$ this is particularly challenging numerically,
requiring e.g. normalization constants which do not have close-form expressions
\cite{DBLP:conf/iclr/YinW0C23, DBLP:conf/iclr/GilitschenskiSS20, DBLP:journals/corr/abs-2006-04437, DBLP:conf/nips/MohlinSB20, DBLP:conf/rss/PeretroukhinGGR20}.
Here, this will be addressed by a tangent space formulation.

In short, the contributions of this paper are:
\begin{itemize}[itemsep=0pt,topsep=-3pt]
\item A HPE model yielding improved accuracy over the current state of the art.
\item Introduction of extended training data, including ablation studies.
\item Novel multitask head/loss and data augmentation designs.
\item A novel tangent-space approach for rotational uncertainty estimation.
\end{itemize}
While most ingredients are nothing new, their combination into competitive designs
in the field of HPE is. The model was integrated into \censored{the FOSS OpenTrack}
\footnote{\censored{\url{https://github.com/opentrack/opentrack}}}, and will therefore be
referred to as OpNet. The full source code for training and evaluation is available on
github \footnote{\censored{\url{https://github.com/opentrack/neuralnet-tracker-traincode/tree/paper}}}.

\section{Methods}

\subsection{Model Design}

\newcommand{\smoothclip}{\mathit{smoothclip}}

The model architecture consists of a feature extraction backbone, global pooling, $50 \%$ dropout, 
and a linear layer per prediction head. The raw features of each head are mapped to final predicted quantities, respectively.
As input, the model takes a $129 \times 129$ monochrome face crop. The motivation for monochrome inputs is invariance 
to hue changes in the local illumination conditions.
ResNet18 \cite{DBLP:conf/cvpr/HeZRS16} and MobileNetV1 \cite{DBLP:journals/corr/HowardZCKWWAA17} were picked as feature
extractors since they are lightweight and proved to work well on various tasks.
To enable full 6 DoF tracking, and to enable learning from landmark-only annotations, the model has the following outputs:
3D rotation, 2D position and size, facial shape parameters, bounding box, and uncertainty parameters.

\textbf{Rotations} are represented by quaternions, which are suitable for limited range HPE \cite{COBO2024110263},
avoiding the gimbal-lock problem of Euler angles. Note that quaternions $q$ and $-q$ represent the same orientation. 
Accuracy was better if this ambiguity was avoided and the network was biased toward identity output. Hence, the 
quaternion is formed from the feature $z_{1} \dots z_{4}$
by $q = q^\prime/||q^\prime||, q^\prime = i z_1 + j z_2 + k z_3 + \smoothclip(z_4)$,
where $i$,$j$,$k$ is the imaginary basis, $\smoothclip(x) = \mathit{ELU}(x)+1$ maps $\mathbb{R}$ to $\mathbb{R}^+$,
and $\mathit{ELU}$ is the function introduced in \cite{DBLP:journals/corr/ClevertUH15}.

\textbf{2D position} and \textbf{size} are both estimated in image space, normalized to $[-1,1]$. The 2D position is 
taken identically from the respective linear layer. The size feature is passed through $\smoothclip$ in 
order to guarantee a positive value.

The model outputs \textbf{shape parameters} for the Basel Face Model (BFM) \cite{DBLP:conf/avss/PaysanKARV09} which consists 
of a 3D base-geometry modified by a combination of deformation basis vectors. Following the literature \cite{DBLP:conf/cvpr/ZhuLLSL16,DBLP:journals/pami/ZhuLLL19,
DBLP:conf/eccv/GuoZYYLL20,DBLP:conf/3dim/WuXN21,DBLP:journals/tip/RuanZWWW21}, the parameters are coefficients for 50 bases
to realize different facial shapes and expressions. Only 68 points which make up the \textbf{3D landmarks} are actually
computed. They adhere to the MultiPIE 68-point markup \cite{DBLP:conf/iccvw/MilborrowBN13,DBLP:journals/ivc/SagonasATZP16}.

The rationale for \textbf{bounding box} prediction is practicality, namely tracking the face through a video sequence as 
in the demo from \cite{DBLP:conf/eccv/GuoZYYLL20}. The output is parameterized by center and size, where the size-features
are mapped by $\smoothclip$ to positive values.

\textbf{Rotation uncertainty} is considered in the tangent-space of rotations $\mathfrak{so}(3)$ which essentially encodes
offsets from a particular orientation by rotation vectors. Thus, the data variation around the predicted pose can be
described by a standard multivariate normal distribution. To eliminate redundancy, the center of the distribution is 
fixed at zero. It has been shown that for small variance this formulation approximates a distribution on $SO(3)$ asymptotically
\cite{DBLP:journals/tac/Lee18, DBLP:conf/cdc/Lee18, DBLP:conf/iclr/YinW0C23}.

Consequently, we must parameterize a covariance matrix $\bf \Sigma$. 
To this end, the network outputs a lower triangular matrix $\bf M \in \mathbb{R}^{3 \times 3}$ filled with
six features taken from a BatchNorm (BN) layer added after the final linear layer. Intuitively, BN helps decouple 
the learning of the magnitude of the variance from the influence of unrelated losses. Without it, the networks didn't
perform well. Hence, given $\bf M$, the covariance matrix is set to
$\bf \Sigma = M M^T + \epsilon \bf I$ resembling a Cholesky decomposition, but the diagonals of $\bf M$ do not need to 
be positive and the addition of $\bf I$ scaled with a small constant ensures strict positivity. 
Note that $\bf M M^T$ is symmetric positive semi-definite.

\textbf{Position and size uncertainty} is modeled by a 3D multivariate normal distribution for the triplet combining 2D
position and head size. Its covariance matrix is constructed like in the case of rotations.

\subsection{Losses}
The training procedure minimizes the sum of individual losses, corresponding to the predicted quantities (in the following
marked with hat $\hat{.}$).

In general, mean and variance parameters $\hat{\mu}$ and $\hat{\sigma}$ are learned by minimizing the negative log likelihood (NLL), 
i.e. $-\log p(y | \hat{\mu}(x),\hat{\sigma}(x))$ of the data density $p$ over some dataset consisting of input-output tuples $(x,y)$
\cite{DBLP:conf/nips/Lakshminarayanan17,DBLP:conf/nips/KendallG17}. Naive approaches have been reported 
as unstable \cite{mve_learning_1994,DBLP:journals/ijon/SluijtermanCH24}. As remedy, an initial period where the 
variance is fixed was suggested. For historical reasons the present model is instead trained with a combination of traditional 
regression losses and NLL, and for "symmetry" reasons NLL losses are employed even for shape parameters, bounding box and landmarks, in 
case of which variances are learned as auxiliary parameters independent of the inputs. 

\newcommand{\shp}{\phi}
\newcommand{\lmk}{\xi}
\newcommand{\nll}{\textit{NLL}}

For the \textbf{rotation} prediction we penalize the geodesic distance by the losses
\begin{align}
    L_{rot}(\hat{q},q) & = 1 - |\hat{q} \cdot q|^2  \\
    \nll_{rot}(\hat{q},q,{\bf \hat{\Sigma}}_{rot}) & = -\log f(\textit{Im} \log(\hat{q}^{-1} q) | 0, {\bf \hat{\Sigma}}_{rot}) \text{,}
\end{align}
where $L_{rot}$ was inspired by the metrics surveyed in \cite{DBLP:journals/jmiv/Huynh09}. The $\cdot$ signifies the inner product of vectors.
In $\nll_{rot}$, $f$ denotes the probability density, $\log$ returns an imaginary quaternion containing the rotation vector,
from which $\textit{Im}$ extracts the imaginary part as 3d vector, the length of which is the geodesic distance, i.e.
the smallest rotation magnitude between $\hat{q}$ and $q$.

Furthermore, considering \textbf{position and size} stacked in a 3d vector $p$, the employed losses are L2 and NLL with 
the normal distribution with variance ${ \bf \hat{\Sigma}}_p$. We thus define
\begin{align}
    L_{p}(\hat{p},p) & = ||p - \hat{p}||^2 \label{eqn:pos_size_loss} \\
    \nll_{p}(\hat{p},p,{\bf \Sigma}_{p}) &= -\log f(p | \hat{p}, {\bf \Sigma}_{p}) \text{.} \label{eqn:pos_size_nll}
\end{align}

The \textbf{shape parameters}, denoted $\shp_i$, are assumed to be distributed independently normal. This simplifies the 
covariance to a diagonal matrix ${\bf \hat{\Sigma}}_{shp} = \textit{diag}(\sigma_1, \sigma_2, \dots)$. 
The corresponding losses are the L2 loss 
$L_{\shp}(\hat{\shp},\shp)$ and NLL with normal distribution $\nll_{\shp}(\hat{\shp},\shp,{\bf \hat{\Sigma}}_{\shp})$
which are defined analogously to \cref{eqn:pos_size_loss} and \cref{eqn:pos_size_nll}, and omitted for brevity.

For \textbf{landmarks} the L1 loss worked well. The corresponding NLL is based on the Laplace distribution. 
Again, statistical independence is assumed. Then the total loss decomposes into sums over coordinate-wise contributions. 
Moreover, it is useful to apply weights $w_i$ to different parts of the face. Thus, the losses are defined by
\begin{align}
    L_{\lmk}(\hat{\lmk},\lmk) & = \sum_i w_i |\lmk_i - \hat{\lmk}_i| \label{eqn:lmk_loss} \\
    \nll_{\lmk}(\hat{\lmk},\lmk,{\bf \hat{\Sigma}}_{\lmk}) &= - \sum_i w_i \log f_\mathit{Laplace}(\lmk_i | \hat{\lmk}_i, \hat{\sigma}_{\lmk,i}) \text{,} \label{eqn:lmk_nll}
\end{align}
where $i$ runs over the 68 $\times$ 3 spatial landmark coordinates. If only x and y coordinates are available, then the 
summation runs only over those.

The \textbf{bounding box} is trained like the shape parameters using L2 loss $L_{bb}$ and $\nll_{bb}$. Input
to those losses are the box corner coordinates, assuming independence.

In order to encourage the network to output nearly unit quaternions, the term $L_{norm}(\hat{q}^\prime) = |1 - |\hat{q}^\prime||^2$ is added, where $\hat{q}^\prime$
is the unnormalized quaternion.

At last we can define the total loss $L$ by
\begin{align}
           L = & L^\prime + \beta_{total} \nll_{total} \\ 
    L^\prime = & \alpha_{rot} L_{rot} + \alpha_p L_p + \alpha_\shp L_\shp \\
             + & \alpha_\lmk L_\lmk + \alpha_{bb} L_{bb} + \alpha_{norm} L_{norm} \\
    \nll_{total} = &\beta_{rot} \nll_{rot} + \beta_{p} \nll_{p} + \beta_\shp \nll_\shp \\
                            + & \beta_\lmk \nll_\lmk + \beta_{bb} \nll_{bb} \text{,}
\end{align}
with weighting factors $\alpha_{.}$ and $\beta_{.}$.

\subsection{Augmentations}
\label{sec:augmentations}
Data samples consist of the input image, a 2D facial bounding box (BB), and the remaining labels.
They are first subject to geometric transformations where also cropping to the face area is performed.
In principle, a square region of interest (ROI) is generated initially from the BB extending its shortest side. This ROI is
subsequently scaled, rotated, and translated by random amounts. Resampling this (rotated) square at the input resolution creates the
face crop. Additionally, the crop is mirrored with probability $p=1/2$ and rotated by $90\deg$ with $p=1/100$. 
Note that no stretching occurs. Finally, the BB is regenerated by taking the BB around the transformed corner points.

Afterwards image intensity augmentations are applied \footnote{Using the Kornia package \url{https://kornia.github.io/}}.
This process is designed to occasionally produce strong distortions with the intent to facilitate generalization to overexposure,
noisy low-light images and similar challenging inputs. First, from Equalize, Posterize, Gamma, Contrast, Brightness, and Blur, 
four are picked and applied with small probabilities up to $p = 1/10$ chance. Then Gaussian noise is randomly added with 
$p \approx 1/2$ and scale up to $\sigma=16/256$ w.r.t. 
the normalized image intensity.

\section{Datasets}
\label{sec:datasets}

Initially training was conducted on 300W-LP, however, performance turned out lacking and facial-expression dependent systematic
pose errors were noticeable. Hypothetically, lacking diversity in 300W-LP might contribute to that, i.e. limited pitch range,
uniform illumination and mostly open eyes.
This motivated the creation of a new dataset, closely following the creation of 300W-LP which should address these shortcomings.
Thereby only the out-of-plane rotation synthesis was performed using the original 300W-LP source images with their BFM parameters. 

\subsection{Extended 300W-LP Reproduction}
\label{sec:300wlp_expansion}

Let's consider the key ingredients. Firstly, the 3D face model. On the basis of same facial region of the BFM as in 
\cite{DBLP:conf/eccv/GuoZYYLL20}, a smooth transition to a background plane was modeled. The resulting mesh is shown
in the Appendix \cref{fig:300wlp_repro_mesh}. The deformation basis was extended to the new vertices by copying the 
vectors from the closest BFM vertex and attenuating them by a distance-dependent decay factor. To the original basis,
new shapes with closed eyes were added. Given an input image, a depth profile is imposed on the background plane 
according to a monocular depth estimate. This is performed by an off-the-shelf MiDaS model
\footnote{MiDaS v3 - Hybrid, \url{https://pytorch.org/hub/intelisl\_midas\_v2/}} \cite{Ranftl2020,Ranftl2021}.
Then, as in \cite{DBLP:conf/cvpr/ZhuLLSL16}, a new image is generated by projecting the original image onto the 3D mesh, 
rotating the face together with the left or right half of the background, smoothly blending between transformed and 
pristine parts, and rendering the result. In addition to unlit renderings, some faces are lit from the side 
with probability $1/1000$. Closed eyes are sampled with probability $1/2$.

The Appendix contains comparisons with 300W-LP in \cref{fig:300wlp_repro_collage}. Furthermore, scatter
plots of rotation distributions are depicted in \cref{fig:rotation_distribution_300wlp}. 
It shall be said that while useful, the novel eye and illumination additions are far from perfect. The eye 
regions suffer from small misalignment errors and the illumination 
suffers from shadow-mapping artifacts and generally does not look particularly realistic. The source code is available in a 
separate repository \footnote{\censored{\url{https://github.com/DaWelter/face-3d-rotation-augmentation}}}.

\subsection{WFLW \& LaPa Large Pose Extension}
\label{sec:wflw_lapa_large_pose}

This section covers further expansion of the training data to in-the-wild 
images where no pose annotations are available. As an improvised solution, 2D landmark annotations 
were leveraged. Perfect labels are thereby not the goal but that the network could learn from relative differences between
frames generated by out-of-plane rotations synthesis.

Inspired by the face-alignment methods 
\cite{DECA:Siggraph2021,Zielonka2022TowardsMR,Zielonka2022TowardsMR,EMOCA:CVPR:2021} (based on the FLAME head-model \cite{FLAME:SiggraphAsia2017}), 
the general idea is to fit 3D landmarks of the BFM to the 2D annotations. Those methods incorporate also photometric fitting and other techniques.
Here, to keep things simple and consistent with 2D, only the visible side of the BFM is used. Then indeed landmarks alone are not sufficient to 
identify plausible 3D reconstructions. To remedy this, initial guesses and pose priors were obtained from a small neural network ensemble trained without the extended data,
and the space of possible shape parameters was soft-constrained by incorporating a NLL loss of a Gaussian mixture
which was fitted to the shape parameters in the 300W-LP dataset.
Ultimately, the labeling process consists of solving a standard 
minimization problem for the sum of several losses: landmark error, rotation error from the prior, 
the shape NLL, as well as soft-constraints for quaternion normalization and non-negative head-size.
The result was manually curated, removing poorly fitted frames. Afterwards the rotation expansion 
from \cref{sec:300wlp_expansion} was applied. Code and notebooks to reproduce every step is available in the source repository.

The procedure was applied to WFLW \cite{DBLP:conf/cvpr/Wu0YWC018} and LaPa \cite{DBLP:conf/aaai/LiuSSSWM20}.
Appealing for this paper, they consist of facial images of a large variety of individuals, poses, and occlusions. 
WFLW comprises $7.5\cdot10^3$ training images with 98 manually annotated
landmarks. The landmarks were converted to 68 points by interpolation. LaPa contains ca. $1.8\cdot10^4$ images annotated with
106 points. However, the latter includes images from 300W-LP which were excluded due to the overlap. The remaining 
images are from Megaface \cite{DBLP:conf/cvpr/Kemelmacher-Shlizerman16}. Ultimately, there were $4942$ images from LaPa 
expanded to ca. $7.7\cdot10^4$, and $1554$ images from WFLW expanded to ca. $2.2\cdot10^4$.
The created datasets are provided online.
\Cref{fig:lapa_extension_collage} and \cref{fig:wflw_extension_collage} in the Appendix show sample images.

\subsection{Face Synthetics}
The Face Synthetics (FS) dataset \cite{wood2021fake} consists of $10^5$ fully synthetic, photorealistic, 
rendered human heads, annotated with segmentation masks and 3D landmarks. The subjects are composed of randomly sampled face shape, hairstyle, 
accessories, skin color, superimposed on a variety of backgrounds. Thus, the annotations are perfect and artifacts
from 300W-LP-style out-of-plane rotations are absent. The authors provide only the 3D landmarks and of 
those only the x and y coordinates. Therefore, only the corresponding landmark losses $L_{\lmk}$ and $\nll_{\lmk}$ are enabled.
The facial bounding box is constructed based on the segmentation, encompassing 
pixels marked as "face" and "nose". Samples where a side length is less than 32 pixels are filtered out.

\section{Implementation Details}
\label{sec:details}

The network is trained with the ADAM optimizer \cite{DBLP:journals/corr/KingmaB14} for $N = 15 $M samples with a maximum
learning (LR) rate of $10^{-3}$. The LR ramps up for $N/20$ samples and decreases after $N/2$ samples to $1/10$th. 
After $2/3 N$ samples, stochastic weight averaging \cite{DBLP:conf/uai/IzmailovPGVW18} is enabled. Furthermore, gradient
clipping with threshold $0.1$ is used. Training time is a few hours on standard desktop hardware.

Facial BB's are often taken around the annotated landmarks \cite{DBLP:journals/tip/RuanZWWW21,DBLP:conf/eccv/GuoZYYLL20}. 
Here, they encompass the full reconstructed facial section of the BFM taken from \cite{DBLP:conf/eccv/GuoZYYLL20} (using all of its vertices).
In case of Biwi, which provides neither landmarks nor boxes, boxes are extracted from the annotations file
provided with \cite{COBO2024110263} and shrunk by $80\%$.
As a result, the facial BB's are consistent across the 300W-LP family, AFLW2000-3D, Biwi and FS.

Regarding cropping, the scale factor determining the facial ROI is sampled from $\mathcal{N}(s,0.1)$, with $s=1.1$ and subsequently clipped to $[s-0.5,s+0.5]$.
Next, consider the ROI translation after scaling. A maximum movement of $t = \tfrac{1}{2}\max(0,\mathit{roi}-\mathit{bb}) + \tfrac{1}{3}\mathit{bb}$ is allowed,
where $\mathit{roi}$ and $\mathit{bb}$ stand in sloppy notation for the extent of the expanded ROI and the original box's side length, respectively. The concrete
translation is sampled from $\mathcal{N}(0,t/2)$ and clipped to $[-t,t]$. This design allows some translation for zoomed-in crops and otherwise placement
of the face anywhere in the crop such that $70 \%$ of the BB remains visible.
The ROI rotation angle is sampled from a uniform distribution between $-30$ and $30\deg$.

As intensity augmentation, noise is in fact potentially added twice, once with probability $p=1/2$ and $\sigma=4$ and secondly with $p=0.1$ 
and $\sigma=16$. This redundant application is implemented as such purely for convenience.

Multiple datasets are combined via
simple random draws. First a dataset is picked with a certain frequency, followed by picking a sample from the dataset
with replacement.

Regarding the landmark weights, eye centers (i.e. top and bottom, 8 point in total) are weighted with $w_i=0$ since good 
samples with closed eyes are scarce.
The loss weights are $\alpha_{rot}=1$,$\alpha_p=1$,$\alpha_\shp=0.01$,$\alpha_\lmk=1$,$\alpha_{bb}=0.01$,
$\alpha_{norm}=10^{-6}$, $\beta_{total}=0.01$ to bring the NLL range to the same order of magnitude 
as the other losses, $\beta_{rot}=1.$,$\beta_p=1.$,$\beta_\shp=0.01$,$\beta_\lmk=1$ and $\beta_{bb}=0.01$.

\section{Results}
\label{sec:result}

The model was evaluated on AFLW2000-3D and Biwi introduced in \cref{sec:intro}. Furthermore, some ablation experiments
were conducted as well as an analysis of the noise response and effectiveness of the uncertainty estimation.

Evaluations were performed on five different networks and the metrics were averaged. 
Reported are the absolute errors of Euler angles, the mean 
of those (MAE), as well as the average of the geodesic errors ($||\textit{Im} \log(\hat{q}^{-1} q)||$).
The largest observed standard error of the sample mean was $0.03\deg$ among all evaluations of AFLW2000-3D and
$0.07\deg$ for Biwi.

The baseline (BL) was trained on the combination of custom large pose expansions of 300W-LP, WFLW and LaPa 
with sampling probabilities $50\%$, $33\%$ and $16\%$, 
respectively. These frequencies were picked ad-hoc, guided by the size of the datasets, preliminary experiments, 
and the quality of the BFM fits. Optimizing them is left as potential future work.
Later on Face Synthetics was added (BL + FS), using the frequencies $50\%$ 300W-LP, $33.3\%$ WFLW, 
$8.4\%$ LaPa, and $8.4\%$ FS. The low amount of FS samples was motivated by its fully synthetic nature and incomplete labels.

A perhaps notable aspect in this work is the consistent avoidance of stretching the input faces. For Biwi at least,
it is common to extract and resize the area under the facial BB to the input size \cite{DBLP:conf/cvpr/YangCLC19, COBO2024110263}.
Instead, consistent with training, the BB is first expanded to a square (and enlarged by $10\%$).

In the evaluation of AFLW2000-3D, the standard protocol in \cite{DBLP:conf/cvpr/RuizCR18}
was followed apart from the input crop, including the removal of 30 samples 
with yaw, pitch or roll angles larger than $99\deg$.
\Cref{table:rot_mae_aflw2k3d} shows a comparison with literature values. Evidently, the 
BL is already very accurate, yet adding FS, yielded further improvement
from $3.19$ to $3.15\deg$ MAE, improving over SOTA by over two sigmas.

The benchmarking on Biwi, the results of which are presented in \cref{table:biwi_results}, follows the
experimental protocol from \cite{DBLP:conf/cvpr/YangCLC19} (apart from the crop), and uses exactly the same frames and facial BB's as in
\cite{COBO2024110263}. Evaluations with the alignment strategy from \cite{COBO2024110263} were also conducted, compensating for different camera angles
and other biases between coordinate systems. Without it, results are modest. With alignment, the accuracy improves drastically. 
However, results for 6DRepNet are not readily available, and a re-evaluation was out-of-scope. Whether aligned results are representative
of the true performance can be questioned because as per \cite{COBO2024110263} the alignment is performed on a per-sequence/individual basis, thus eliminating
systematic biases caused by the subject's appearance.

\begin{table}[h]
    \caption{Rotation errors in degrees from the AFLW2000-3D benchmark. YPR stands for yaw,pitch and roll, MAE for their average and Geo for the geodesic error.}
    \begin{tabular}{lrrrrr}
        Method & Y  & P & R & MAE & Geo. \\
        \hline
        \hline
        HopeNet \cite{DBLP:conf/cvpr/RuizCR18}                   & 6.47 & 6.56 & 5.44 & 6.15 & 9.93 \\
        FSA-Net \cite{DBLP:conf/cvpr/YangCLC19}                  & 4.50 & 6.08 & 4.64 & 5.07 & 8.16 \\
        WHENet \cite{DBLP:conf/bmvc/ZhouG20}                     & 4.44 & 5.75 & 4.31 & 4.83 & -    \\
        TokenHPE \cite{DBLP:conf/cvpr/ZhangLDXL23}               & 4.36 & 5.54 & 4.08 & 4.66 & -    \\
        QuatNet \cite{DBLP:journals/tmm/HsuWWWL19}               & 3.97 & 5.61 & 3.92 & 4.50 & -    \\
        LSR \cite{DBLP:journals/pr/CelestinoMNC23}               & 3.81 & 5.42 & 4.00 & 4.41 & -    \\
        MFDNet \cite{DBLP:journals/tmm/LiuFZLLW22}               & 4.30 & 5.16 & 3.69 & 4.38 & -    \\
        EHPNet \cite{DBLP:conf/icpram/ThaiTBNT22}                & 3.23 & 5.54 & 3.88 & 4.15 & -    \\
        6DRepNet \cite{DBLP:conf/icip/HempelAA22}                & 3.63 & 4.91 & 3.37 & 3.97 & -    \\
        img2pose \cite{DBLP:conf/cvpr/AlbieroC0PH21}             & 3.42 & 5.03 & 3.27 & 3.91 & 6.41 \\
        6DoF-HPE \cite{ALGABRI2024122293}                        & 3.56 & 4.74 & 3.35 & 3.88 & -    \\
        MNN \cite{DBLP:journals/pami/ValleBB21}             & 3.34 & 4.69 & 3.48 & 3.83 & -    \\
        SADRNet \cite{DBLP:journals/tip/RuanZWWW21}              & 3.93 & 5.00 & 3.54 & 3.82 & -    \\
        DAD-3DHeads \cite{DBLP:conf/cvpr/MartyniukKKKMS22}         & 3.08 & 4.76 & 3.15 & 3.66 & -    \\
        SynergyNet \cite{DBLP:conf/3dim/WuXN21}                  & 3.42 & 4.09 & 2.55 & 3.35 & -    \\
        SRHP \cite{COBO2024110263}                               & 2.76 & 4.25 & 2.76 & 3.26 & 5.29 \\
        DSFNet \cite{DBLP:conf/cvpr/LiWCKT23}                    & \bf 2.65 & 4.28 & 2.82 & 3.25 & -    \\
        \hline
        \hline
        \bf OpNet BL       & 2.80 &     4.22 &     2.54 &     3.19 &     5.26 \\
        \bf OpNet BL + FS  & 2.79 & \bf 4.18 & \bf 2.49 & \bf 3.15 & \bf 5.23
    \end{tabular}
    \label{table:rot_mae_aflw2k3d}
\end{table}

\begin{table}[h]
    \caption{Rotation errors from the Biwi benchmark.}
    \begin{tabular}{lcrrrrr}
        Method & Y  & P & R & MAE & Geo. \\
        \hline
        \hline
        FSA-Net \cite{DBLP:conf/cvpr/YangCLC19}           &    4.27 &     4.96 &      2.76 &     4.00 & 7.64 \\
        SRHP (6D) \cite{COBO2024110263}                   &    4.58 &     4.65 &      2.71 &     3.98 & 7.30  \\
        img2pose \cite{DBLP:conf/cvpr/AlbieroC0PH21}      &    4.56 &     3.54 &      3.24 &     3.78 & 7.10 \\
        HopeNet \cite{DBLP:conf/cvpr/RuizCR18}            &    4.53 & \bf 3.08 &      2.83 &     3.48 & 6.60 \\
        WHENet \cite{DBLP:conf/bmvc/ZhouG20}              &    3.60 &     4.10 &      2.73 &     3.48 & - \\
        6DRepNet \cite{DBLP:conf/icip/HempelAA22}         &\bf 3.24 &     4.48 &      2.68 &     3.47 & - \\
        6DRepNet360 \cite{6drepnet360}                    &    3.37 &     3.87 &      2.93 & \bf 3.39 & - \\
        \hline
        \bf OpNet BL                                      &     3.80 &     4.93 &     2.57  &    3.77  & 7.21 \\
        \bf OpNet BL + FS                                 &     3.66 &     4.61 & \bf 2.44  &    3.57  & 7.01 \\
        \hline
        \multicolumn{6}{c}{Aligned} \\
        \hline
        HopeNet \cite{DBLP:conf/cvpr/RuizCR18}            &     4.53 &     3.08 &      2.83 & 3.48      &     6.60 \\
        img2pose \cite{DBLP:conf/cvpr/AlbieroC0PH21}      &     4.04 &     3.12 &      3.03 & 3.40      &     6.23 \\
        SRHP (Euler) \cite{COBO2024110263}                &     3.98 &     3.09 &      2.40 & 3.16      &     5.42  \\
        FSA-Net \cite{DBLP:conf/cvpr/YangCLC19}           &     3.59 &     2.90 &      2.27 & 2.92      &     5.36 \\
        \hline
        \bf OpNet BL                                      &     2.84 &     2.75 &      2.90 &     2.83  &     5.01 \\
        \bf OpNet BL + FS                                 &     2.57 &     2.47 &      2.92 &     2.65  &     4.72
    \end{tabular}
    \label{table:biwi_results}
\end{table}

\begin{figure*}[h]
    \centering
    \includegraphics[width=1.0\linewidth]{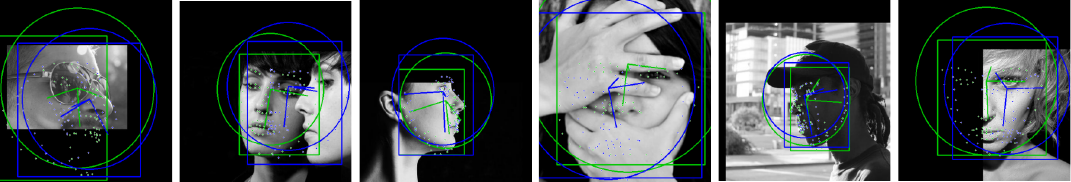}
    \caption{Visualization of the samples from AFLW2000-3D with the worst rotation error. Predictions are blue, ground truth is green.
    Shown are axes of the local coordinate system, landmarks and bounding box.}
    \label{fig:worst_fits}
\end{figure*}

\begin{figure*}[h]
    \centering
    \includegraphics[width=1.0\linewidth]{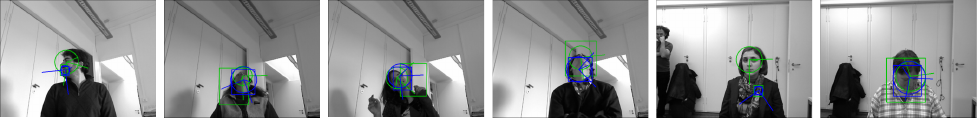}
    \caption{Visualization of the samples from Biwi with the worst rotation error, analogous to \cref{fig:worst_fits}}
    \label{fig:worst_fits_biwi}
\end{figure*}

\Cref{fig:worst_fits} visualizes the worst estimates judging by rotation error. It reveals failures in 
situations with heavy occlusion and a sample with two visible faces. It also shows apparently mislabeled samples
where the predictions look more plausible.

\Cref{table:ablation_method} shows results from various method ablations. Increasing the backbone capacity from 
MobileNet to ResNet18 improved the metrics only insignificantly. Removal of landmark predictions yielded a small improvement
in MAE over the BL. Hypothetically, synergetic effects between tasks did not occur in the BL and now more capacity was
freed for pose prediction. However, the landmark prediction was needed to utilize the annotations of the Face Synthetics data. 
The variant without landmarks and ResNet backbone would have been very strong if only geodesic distance was considered but the MAE metric suffered 
so drastically that it cannot be considered the best. The other modifications worsened the accuracy in both metrics. 
Interestingly, in-plane rotation augmentation had a big impact, where a smaller rotation range yielded intermediate results.

\begin{table}[h]
    \caption{Ablation study with different variations of the methodology. Every line means a change from the baseline (OpNet BL). 
    Changes are not cumulative with other lines. ResNet18 means the feature extractor was replaced
    with it. A minus means removal of the ingredient. "Intensity Aug." refers to the image intensity augmentations, 
    "NLL" to the NLL losses, "Landmarks" to the landmark losses, "In-plane Rot." to the respective rotation augmentation,
    and "$5\deg$ In-plane Rot." to overridden rotation limits.}
    \begin{tabular}{lrr}
    Method Variation     & MAE  & Geo. \\
    \hline
    \hline
    OpNet BL               &      3.19 &     5.26 \\
    \hline
    ResNet18               &      3.18 &     5.24 \\
    - Landmarks            &  \bf 3.16 &     5.26 \\
    ResNet18 - Landmarks   &      3.27 & \bf 5.21 \\
    - Intensity Aug.       &      3.24 &     5.37 \\
    - NLL                  &      3.30 &     5.35 \\
    - In-plane Rot.        &      3.53 &     5.65 \\
    $5\deg$ In-plane Rot.  &      3.44 &     5.56 \\
    \end{tabular}
    \label{table:ablation_method}
\end{table}

\Cref{table:ablation_data} shows an ablation study for dataset variations. Starting from modest results with
300W-LP, the accuracy improves as more data is added.
Interestingly, even the basic 300W-LP reproduction (R-300W-LP) improved performance. The reason is unknown, but it could be explained by a 
slightly different rotation distribution or the 3D geometry in particular due to the depth estimation. 
Adding directional lighting and closed eyes yielded a further boost.

\begin{table}[ht]
    \caption{Ablation study with different training distributions. "300W-LP" refers to the original dataset, "FS" to
    Face Synthetics, "R-300W-LP" to the reproduction, "RA-300W-LP" the reproduction with closed eyes and profile 
    illumination, and finally "EX" to the extension generated from LaPa and WFLW.}
    \begin{tabular}{rlrr}
    Notes & Dataset              & MAE  & Geo. \\
    \hline
    \hline
    & 300W-LP              &      3.44 &     5.44  \\
    & 300W-LP + FS         &      3.34 &     5.36  \\
    & R-300W-LP            &      3.28 &     5.34  \\
    & RA-300W-LP           &      3.27 &     5.27  \\
    \hline
    BL      & RA-300W-LP + EX &   3.19 &     5.26  \\
    BL + FS & RA-300W-LP + EX + FS &  \bf 3.15 & \bf 5.23  \\
    \end{tabular}
    \label{table:ablation_data}
\end{table}

\begin{figure}[ht]
    \centering
    \includegraphics[width=0.9\linewidth]{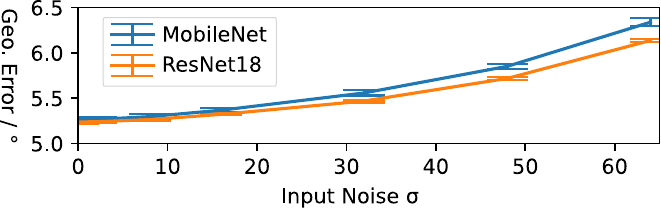}
    \caption{Plots the geodesic error of rotation predictions versus the standard deviation $\sigma$ of Gaussian noise added to input images.
    The evaluations are conducted over AFLW2000-3D modified by noise. The error bars show the standard error of the sample mean 
    over the five evaluation networks.
    }
    \label{fig:noise_resist}
\end{figure}

\begin{figure}[ht]
    \centering
    \includegraphics[width=0.55\linewidth]{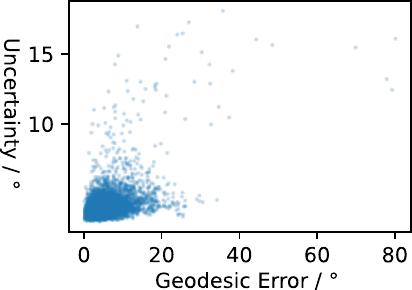}
    \caption{Correlation of the uncertainty estimate with rotation errors. The data points iterate over samples from
    AFLW2000 and the five BL evaluation networks. Recall that the uncertainty estimate ${\bf \hat{\Sigma}}_{rot}$ is a covariance matrix. 
    Plotted is its Frobenius norm to condense it to a single number.}
    \label{fig:uncertainty}
\end{figure}

Aside of benchmark outcomes, in practice the ResNet18 variant produced subjectively noticeably 
smoother predictions than the BL. This motivated an evaluation on noisy inputs, the results of which are shown in 
\Cref{fig:noise_resist}. And indeed, as the noise is increased, the gap between
rotation errors widens to ca $0.25\deg$ which might explain the subjective feeling. On the other hand this gap amounts to
only $5\%$ of the total error magnitude, so other aspects could play a role.

\Cref{fig:uncertainty} demonstrates some degree of effectiveness of the uncertainty estimation. Note that the 
failure cases with errors larger than $40\deg$ are correctly attributed with correspondingly large uncertainty.
Overall the correlation between pose error and uncertainty is rather weak.

Since the network is equipped with a landmark prediction head, it was also evaluated on the AFLW2000-3D sparse
face alignment benchmark following the protocol in \cite{DBLP:conf/cvpr/ZhuLLSL16}.
The metric thereof measures the distance of the 68 3D-landmarks from ground truth labels using the normalized mean
absolute error of the x and y components (NME 2D), ignoring the depth coordinate. It is
computed separately for three yaw bins, $\left[0\deg,30\deg\right)$, 
$\left[30\deg,60\deg\right)$ and $\left[60\deg,90\deg\right)$. Respective results are presented in \cref{table:landmark_nme2d} 
together with literature values from prior art. As can be seen, the accuracy is decent but not up to current SOTA. However, OpNet
was also not optimized for landmark prediction.

\begin{table}[h]
    \caption{NME 2D evaluation on AFLW2000-3D. The middle three columns correspond to the yaw bins and "Mean" shows the average of the three.}
    \begin{tabular}{lrrrrr}
        Method             & 0-30 & 30-60 & 60-90 & Mean \\
        \hline
        \hline
        3DDFA \cite{DBLP:conf/cvpr/ZhuLLSL16} & 3.78 & 4.54 & 7.93 & 5.42 \\
        PRNet \cite{DBLP:conf/eccv/FengWSWZ18} & 2.75 & 3.51 & 4.61 & 3.62 \\
        3DDFA V2 \cite{DBLP:conf/eccv/GuoZYYLL20} & 2.63 & 3.42 & 4.48 & 3.51 \\
        SADRNet \cite{DBLP:journals/tip/RuanZWWW21} & 2.66 & 3.30 & 4.42 & 3.46 \\
        SynergyNet \cite{DBLP:conf/3dim/WuXN21} & \bf 2.65 & 3.30 & 4.27 & 3.41 \\
        JVCR  \cite{zhang2019adversarial} & 2.69 & \bf 3.08 & \bf 4.15 & \bf 3.31 \\
        \hline
        \hline
        OpNet BL       & 2.80 &  3.54 & 4.43 &  3.59 & \\
        OpNet BL + FS  & 2.75 &  3.48 & 4.41 &  3.55 &
    \end{tabular}
    \label{table:landmark_nme2d}
\end{table}

\section{Conclusion}

This work presents an approach to HPE which achieves highly competitive performance by leveraging 
data and data-augmentation strategies to much greater extend than before.
While none of the ideas are new, we can recognize the degree of their effectiveness.

The results suggest that out-of-plane rotation synthesis from \cite{DBLP:conf/cvpr/ZhuLLSL16} has not yet reached its limit, i.e. 
when slightly improved and applied to a sufficiently large and diverse data volume, significantly better performing models might be trained 
than with the original 300W-LP dataset. Adding a different flavor of synthetic data, namely the Face Synthetics can
boost performance further, where potentially the different image style helps to overcome the domain gap to the real world. 
The extent of improvement is surprising since an otherwise moderately performing
model is boosted to beyond SOTA with quite some margin.

On the other hand the approach in this work was not very effective for face alignment. The paper also does not address the question
of how prior HPE art (which performs better with the 300W-LP baseline) would benefit from the suggested training data.
In the latter regard, the paper is limited in scope. However, re-training models of prior art would be an interesting
direction for future work as it could yield further improvements for HPE and face alignment.
Another direction to pursue would be the acquisition of a synthetic dataset like Face Synthetics but with 
perfect 6 DoF pose labels, abolishing the need to measure the error indirectly via landmark predictions.

As a cautionary tale, the fact that the ResNets lower noise sensitivity only showed when modifying the test set, highlights
the risk of "overfitting" methods to a particular test set - in this case with only clean images.

A merit of this work is also the uncertainty estimation with its tangent-space Gaussian formulation which is straight forward
to implement, provided an accuracy boost when added to the model, and was effective at detecting failure cases.

On the practical side, both the MobileNet and ResNet variants are accurate, efficient models suitable for real-time applications.

{
    \small
    \bibliographystyle{ieeenat_fullname}
    \bibliography{references}
}


\cleardoublepage

\appendix

\section{Appendix}

\begin{figure}[h]
  \centering
  \includegraphics[width=0.33\textwidth]{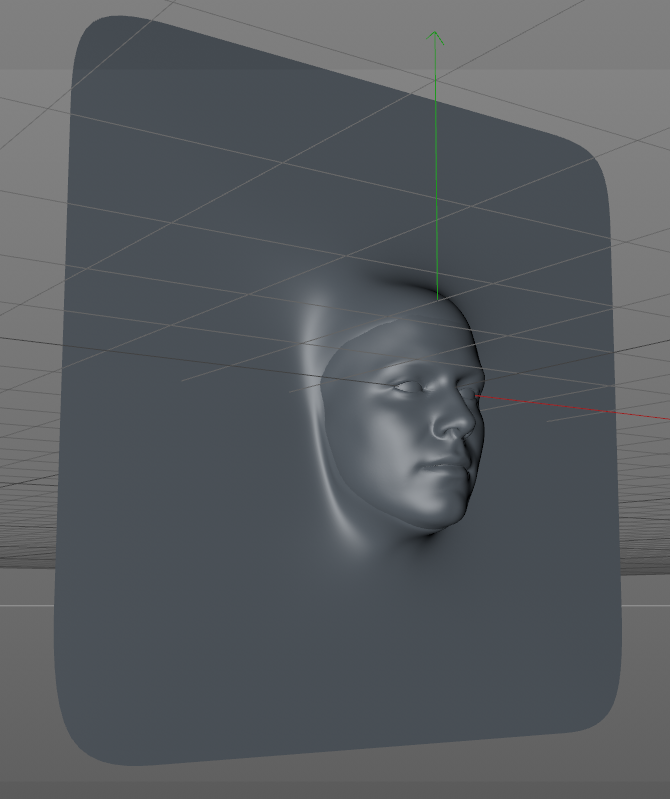}
  \caption{The underlying mesh for rendering images with out-of-plane rotations.}
  \label{fig:300wlp_repro_mesh}
\end{figure}

\begin{figure*}[h]
  \centering
  \includegraphics[width=0.8\textwidth]{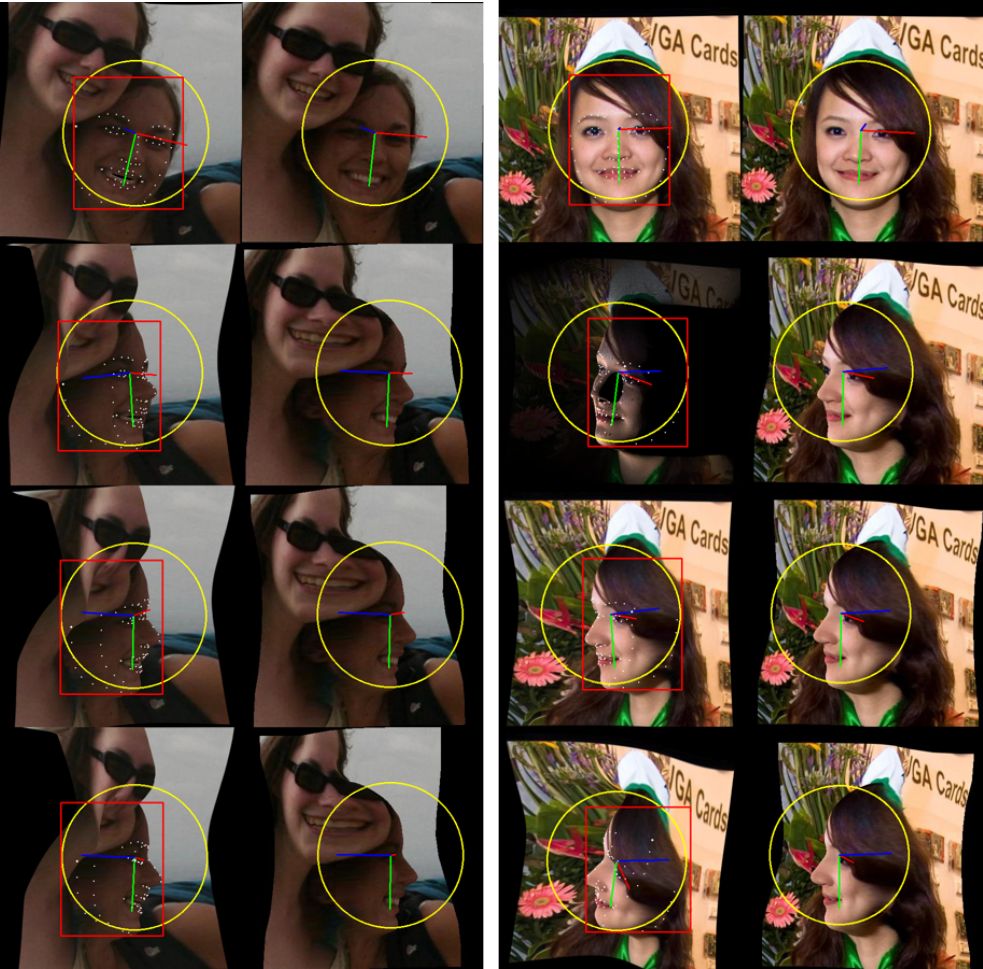}
  \caption{Example frames of the extended 300W-LP dataset. Left and right
  panel show difference subjects. Per panel, left column: new, right column: 300W-LP.}
  \label{fig:300wlp_repro_collage}
\end{figure*}

\begin{figure*}[h]
  \centering
  \includegraphics[width=0.8\textwidth]{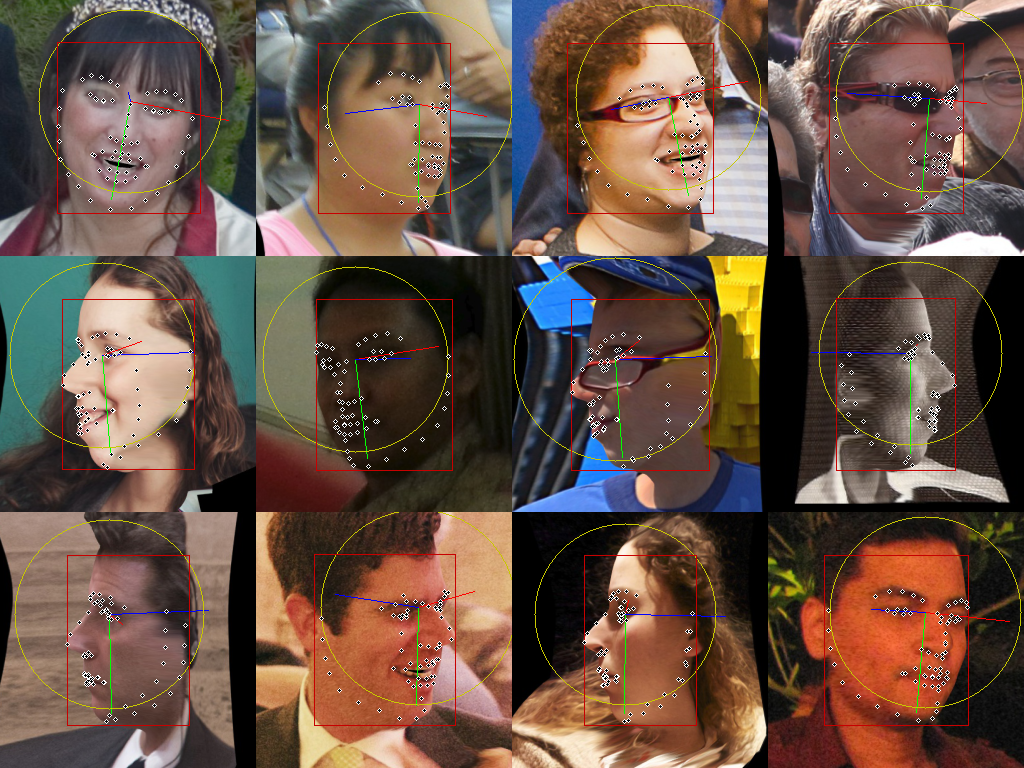}
  \caption{Example frames from the synthetically expanded LaPa dataset.}
  \label{fig:lapa_extension_collage}
\end{figure*}

\begin{figure*}[h]
  \centering
  \includegraphics[width=0.8\textwidth]{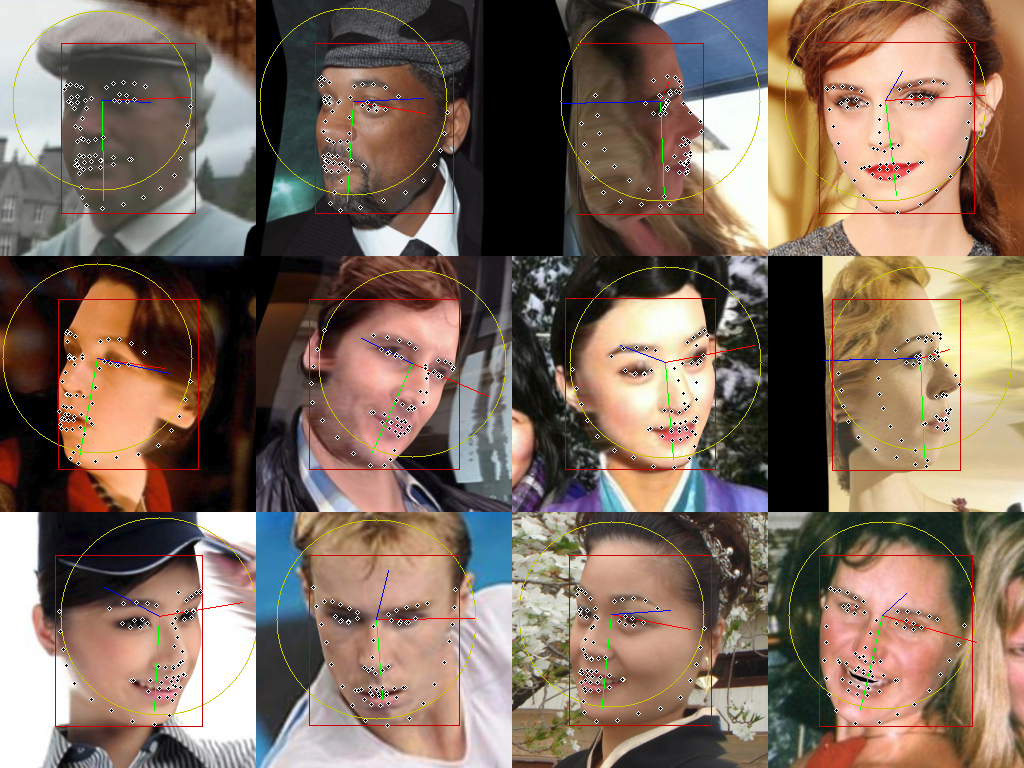}
  \caption{Example frames from the synthetically expanded LaPa dataset.}
  \label{fig:wflw_extension_collage}
\end{figure*}

\begin{figure*}[h]
  \centering
  \includegraphics[width=0.9\textwidth]{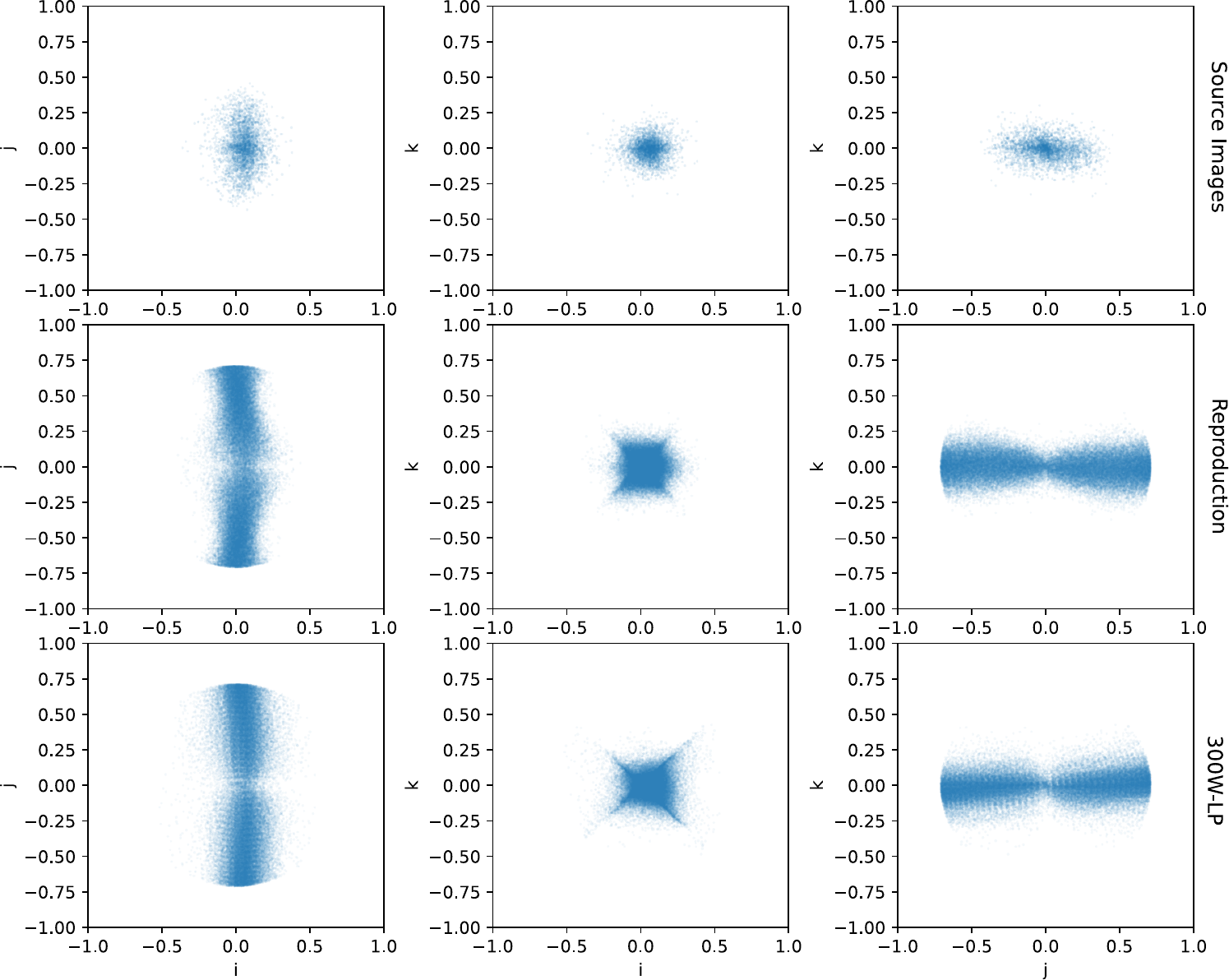}    
  \caption{Rotation distribution of the expansion by in-plane rotations. The plots show projections of the quaternion components
  of the rotation annotations. Every point represents a sample in the respective dataset. Top row: original without expansion.
  Middle: my reproduction. Bottom: 300W-LP.}
  \label{fig:rotation_distribution_300wlp}
\end{figure*}

\end{document}